\pdfoutput=1
\documentclass[10pt,twocolumn,letterpaper]{article}

\usepackage[pagenumbers]{cvpr}

\usepackage[normalem]{ulem}

\usepackage{pifont}
\usepackage{xcolor}

\newcommand{\cmark}{\textcolor{green!60!black}{\ding{51}}} %
\newcommand{\xmark}{\textcolor{red!70!black}{\ding{55}}}   %
\usepackage[T1]{fontenc}

\usepackage{microtype}

\renewcommand{\paragraph}[1]{\vspace{.5em}\noindent\textbf{#1.}}

\usepackage{tabularx}
\usepackage{multirow}
\usepackage{booktabs}
\usepackage{amsmath}
\usepackage{amssymb}

\definecolor{cvprblue}{rgb}{0.21,0.49,0.74}
\usepackage[pagebackref,breaklinks,colorlinks,allcolors=cvprblue]{hyperref}

\usepackage{microtype}

\def\confName{CVPR}
\def\confYear{2027}

\title{RestPose: Canonicalization for Open Object Pose Estimation}

\author{Sandeep Inuganti$^{1,2}$ \and Weihang Li$^{3}$ \and Benjamin Busam$^{3}$
\end{tabular}\\\begin{tabular}[t]{c}
Didier Stricker$^{1,2}$ \and Jason Rambach$^{1}$
\end{tabular}\\\begin{tabular}[t]{c}
$^{1}${\small German Research Center for Artificial Intelligence, DFKI, Germany}\\
$^{2}${\small RPTU Kaiserslautern, Germany} $^{3}${\small Technical University of Munich, Germany}\\
$^{1}${\tt\small \{firstname.lastname\}@dfki.de} $^{3}${\tt\small \{firstname.lastname\}@tum.de}}

\begin{document}
\maketitle

\begin{abstract}
In object pose estimation methods, predicted poses are expressed with respect to a canonical frame, defined by a reference model of that object. In learning-based methods, an arbitrarily chosen modeling convention defines this canonical frame, which is then used to generate the ground-truth pose-annotations for training. Object pose estimation methods are therefore bound to such arbitrary conventions, restricting generalization. In this work, we propose that such a canonical frame can instead be derived for any object, and present RestPose, an approach towards open object pose estimation that does not require reference models or canonical frame conventions during training. From a single input image, we generate an object mesh and canonicalize it through a geometric and physical analysis of the object, then use foundation model-based dense matching to estimate the pose. We choose category-level benchmarks for our experiments, which contain instances across various object categories and enable us to measure the effectiveness of our canonicalization. Across HouseCat6D, NOCS, and Omni6DPose datasets, we improve on the supervised state-of-the-art methods by 18.3 points in absolute metrics and 13.8 points in scale-agnostic metrics, without requiring task-specific training. Code will be published.
\end{abstract}

\section{Introduction}

Estimating the {6D} pose of an object requires a defined canonical frame to measure against. In an instance-level setting~\cite{gigapose,megapose,hipose,jenga}, the frame is defined by the object's {CAD} model. 
Alternatively, when no {CAD} is available, the frame
is often defined from a {reference view}, so that pose is expressed relative to one arbitrary
image of the object~\cite{any6d,pany,oryon}. In a category-level setting, neither
is available at inference - in this setting, every instance of any category is assumed to be seen for the first time during inference, and thereby has no distinguished orientation (canonical frame) until one is arbitrarily chosen.
\begin{figure}[t]
    \centering
    \includegraphics[width=\linewidth]{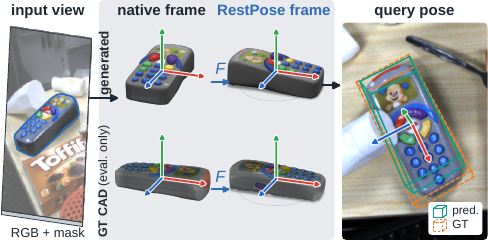}
    \caption{\textbf{Overview.} {RestPose computes a canonical frame from generated geometry.} A masked
    reference view yields a generated mesh of a \emph{remote}. In their native frames (middle) the two
    meshes (generated and GT mesh) lie along different axes.
    RestPose's canonicalization $F$ sets each mesh at rest in a common reference frame - allowing us to design an annotation-free and pose training-free method.
    }
    \label{fig:teaser}
\end{figure}
\begin{figure*}[t]
    \centering
    \begin{minipage}[b]{0.49\textwidth}
        \centering
        \includegraphics[width=\linewidth]{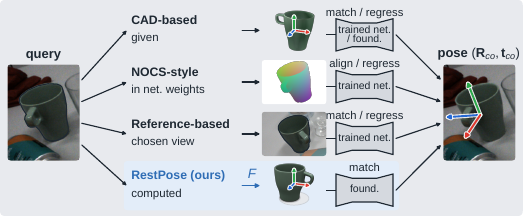}
        \captionof{figure}{{One object instance, four ways of canonicalization.}}
        \label{fig:settings}
    \end{minipage}\hfill
    \begin{minipage}[b]{0.495\textwidth}
    \captionof{table}{{Positioning of RestPose compared to the literature}}
        \centering\footnotesize
        \setlength{\tabcolsep}{2.2pt}
        \newcommand{\hdc}[1]{\begin{tabular}[b]{@{}c@{}}#1\end{tabular}}
        \begin{tabular}{@{}lcccl@{}}
          \toprule
          \textbf{Method type} & \hdc{no pose\\net} & \hdc{no\\CAD} & \hdc{categ.\\frame} & \begin{tabular}[b]{@{}l@{}}source of\\the canonical\end{tabular} \\
          \midrule
          Reference-based~\cite{any6d,pany,onepose}        & \xmark & \cmark & \xmark & the anchor view \\
          CAD-based~\cite{jenga,hipose,megapose} & \xmark & \xmark & \xmark & the given {CAD} \\
          \quad (training-free)~\cite{pos3r,foundpose}       & \cmark & \xmark & \xmark & the given {CAD} \\
           Cat-level NOCS~\cite{nocs,vinet,gcepose}             & \xmark & \cmark & \cmark & convention, in weights \\
          \quad (RGB-only)~\cite{msos,lapose,unidet}          & \xmark & \cmark & \cmark & convention, in weights \\
          \midrule
          \textbf{RestPose (ours)}                         & \cmark & \cmark & \cmark & \textbf{computed} \\
          \bottomrule
        \end{tabular}
        \vspace{4pt}
        
        \label{tab:settings}
    \end{minipage}
    \par\vspace{2pt}
    \begin{minipage}{\textwidth}\small \textbf{Different sources of a canonical frame} in object pose estimation: Model based methods inherit it from the
    object models. NOCS-style methods declare a per category frame and train networks to predict
    object coordinates in it.
    Reference-based methods borrow the frame of an arbitrary view and report pose relative to
    it. RestPose generates the object's
    mesh and derives the frame following a geometric analysis.
    Table column descriptions: no network trained {for pose}; no {CAD} required of
    the test object; a category-level frame shared across instances.
    \end{minipage}
\end{figure*}

A solution to this problem was established by NOCS~\cite{nocs} and was widely adopted by category-level object pose methods~\cite{spd,fsnet,vinet,secondpose,agpose,gcepose,spherepose,spotpose}.
NOCS proposes to {declare} that canonical frame once per category and render coordinate supervision from {CAD}
models aligned into it. A network trained on those renders does not use the category's {CAD} at test time, and it appears to have no dependence on that chosen canonical frame per category.

However, the frame convention remains inherent in the network weights, learned through the pose annotations over every frame and every instance of the training split. In summary, no existing method currently combines being: (1) pose network training-free, (2) GT-reference/CAD model-free, and (3) able to operate on object categories without NOCS. The core obstacle to achieving this combination is the requirement of a canonical frame definition. Instead, defining a canonical frame computation of the objects removes
the above dependencies that bind pose estimation to a fixed set of objects, and provides a pathway to open object pose estimation. RestPose (~\cref{fig:teaser}) motivates this paradigm shift for the object pose estimation domain.

Moreover, we demonstrate how open object pose estimation can leverage foundation
models: A single-image generative model~\cite{sam3d} reconstructs a textured mesh of every object in view and a dense matcher registers correspondences with the
image~\cite{dust3r,mast3r}. 
One can therefore build a {generative object model} of the test instance at inference time, and register against it to predict the pose. This briefly defines our proposed open object pose estimation methodology.

In such a formulation, the canonical frame problem is still present since a generated mesh provides its own arbitrary frame, which follows no specific benchmark convention.
Without canonicalization, such an open object pose estimation pipeline's output is in a frame of the generated object's own, which can be inconsistent.
The alternative natural candidate for a {derived} canonical frame is a physical one which we describe step-by-step in our Methodology (\cref{Sec:method}).
In summary, our main contributions in this work are:
\begin{itemize}\itemsep2pt
    \item We propose \emph{RestPose}, a canonicalization-based pose estimation approach using object's own geometrical and physical properties. This provides an alternative to standard NOCS based approaches, which require {CAD}-based coordinate maps for supervision, rendered with an arbitrary annotation convention.

    \item An 
    annotation-free and pose network training-free pipeline, with object model geometry derived from 3D generator, defining a first step towards open object pose estimation.
    \item We provide a symmetry disambiguation stage using multi-view processing realized with predicted camera poses instead of ground-truth. This stage is post-hoc and optional, requiring no re-inference.
    \item We evaluate on {three benchmarks, across 165 different object categories} that contain numerous object instances. We exceed the supervised state-of-the-art methods, but we also analyze the performance {limit} on what any method built on our canonicalization can achieve.
\end{itemize}

\section{Related Work}

\subsection{Category-Level Pose and the Canonical Frame}

NOCS~\cite{nocs} made category-level pose tractable by defining a canonical frame per category, then
regressing, for each pixel, its coordinates in that frame. Supervision is provided by renders of
{CAD} models aligned into that frame. Subsequent research improved the representation or the
correspondences while inheriting the frame. One line of research deforms a learned category shape
prior~\cite{spd,dpdn}, while another favors direct geometric
reasoning~\cite{fsnet,gpvpose}. Finally a third line reshapes the rotation representation itself:
through decoupled rotations on spherical features~\cite{vinet,spherepose}, {SE(3)}-consistent
feature fusion~\cite{secondpose}, instance-adaptive keypoints~\cite{agpose,spotpose} and global
shape context~\cite{gcepose}.
Generative formulations address the same problem as sampling rather
than regression~\cite{genpose,diff9d}.
Category-level pose benchmarks have also expanded
alongside the methods, from REAL275~\cite{nocs} to cluttered and large-scale category
settings~\cite{housecat6d,omni6dpose}.

The dependence on the canonical frame is rarely discussed, because it is hidden during inference.
The accuracy of the pose estimator is measured against this convention which enters the weights
by frame and instance level training supervision.
We replace the convention with a computed canonical frame and align it to the benchmark's own frames only for quantitative evaluation.

\subsection{Unseen-Object Pose from a Reference Model or a Reference View}

Unseen object pose is agnostic to object categories and generalizes across instances themselves. Some methods assume an object
model at test time and match or render against it~\cite{foundationpose,gigapose,foundpose,megapose,pos3r}.
Others replace the model with reference views of the object~\cite{any6d,pany,oryon,onepose,gspose}.

The field draws this distinction as \emph{model-based} versus \emph{model-free} unseen object pose estimation. Our method's placement pushes the boundaries of this domain. We require no model of the test object, yet
we do register against a generated~\cite{sam3d} object model from the test image itself. However, both these settings do not address the canonical frame issue - they simply define the frame as the one of the object {CAD}. Two objects of the same category can thus receive unrelated
frames, and no generalization about the pose of that category of object is available. 

Among the unseen methods, Pos3R~\cite{pos3r} is the closest to ours during inference: 
It estimates pose by matching the query crop against templates rendered from an object model, with a similar dense matching technique based on MASt3R~\cite{mast3r} that we use. However, it is still a model-based method and therefore, confined to the above discussed limitations.
Any6D~\cite{any6d} is the closest to ours in spirit: reconstructing shape from a
single anchor with an image-to-3D model and fitting size and pose jointly by render-and-compare, yet
it also adopts a given canonical frame.

\subsection{Feed-Forward Geometry Foundation Models}

Where pose pipelines once depended on hand-designed features and minimal solvers~\cite{sift,epnp},
recent models predict dense geometry directly. Two-view methods output pointmaps and
matches~\cite{dust3r,mast3r} while multi-view models output depth, pointmaps and camera poses in one
pass~\cite{vggt,cut3r}. Self-supervised image and point backbones supply the features these build
on~\cite{dinov2,clip,sonata}. These foundational features enable our pipeline, supplying at test time what previously required per-dataset training. In fact, several recent methods  directly build on such features~\cite{pos3r,foundpose,pany}.

\subsection{Canonicalization from Geometry, and its Ambiguities}

Aligning shapes into a common frame without labels is classically attempted with principal axes or detected symmetry. However, both degenerate for the near-symmetric objects which are widely present in object pose benchmarks, e.g. a box has three near-equal principal axes and a bottle a continuous one. An \emph{exact} symmetry is absorbed by the annotation convention or encoded into the correspondence
target~\cite{symcode} used for supervision. However, an exceptionally challenging case is that of a \emph{near} symmetry, which makes the pose well defined while making it visually challenging to distinguish between various views of that object (e.g. visibility of a mug's handle can make it look fully symmetric in some views). We try to address this near-symmetry problem with our multi-view processing.
On the other hand, learned canonical correspondences shift the
convention to a training set~\cite{objectmatch}.
Therefore, a derived frame is a promising
annotation-free alternative for canonicalization.

\paragraph{The RestPose proposal:} We have discussed various strategies which aim to solve object pose estimation in different settings (\cref{fig:settings}). However, unifying these settings and moving towards open object pose estimation is seldom attempted in the literature. A recent work, OPT-Pose~\cite{optpose}, aims to unify absolute and relative pose for unseen objects by training a transformer on the benchmarks we evaluate on. Although it still relies on NOCS-style supervision to achieve its goal, we still take OPT-Pose as our
primary comparison. \Cref{tab:settings} places our method against the above mentioned canonical frame strategies. Our method's placement facilitates an open object pose setting, which future works can explore.

\section{Method}
\label{Sec:method}
\begin{figure*}[t]
    \centering
    \includegraphics[width=\linewidth]{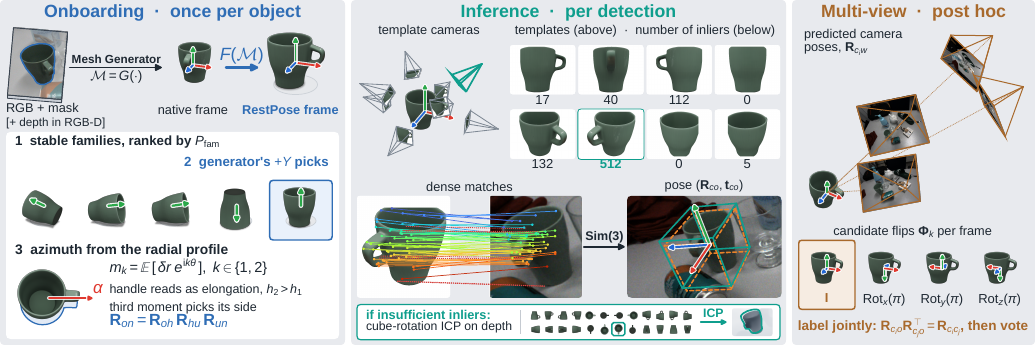}
    \caption{\textbf{RestPose pipeline}. \emph{Onboarding} runs once per object. A
    mesh is generated from one masked view, with its depth in the RGB-D setting, in the generator's native frame. 
    one unique canonical frame is selected,
    and the azimuth is read from the radial profile, giving the canonical rotation.
    \emph{Inference} runs per detection. Templates are rendered from cameras placed
     around the RestPose mesh, and reciprocal matches against each are counted. The template with most inliers wins, and its matches are lifted to a $\mathrm{Sim}(3)$ fit inside RANSAC.
     In case of insufficient inliers, the detection takes the boxed {ICP} branch instead.
    In the RGB-only setting onboarding uses no metric-scale sensor depth and the query's predicted depth has arbitrary scale.
    \emph{Multi-view labelling} runs post hoc on the per-frame outputs. 
    Each frame's estimate is first labelled jointly among the four flips, and the labelled poses then vote for a consensus.
    }
    \label{fig:method}
\end{figure*}

Given a sequence of {RGB} frames $\{I_i\}_{i=1}^{T}$
(and depth map $D_i$ if needed) we predict the object pose $(\mathbf{R}_{c o},\mathbf{t}_{c o})$ in each image's camera frame $c$ w.r.t object's RestPose frame $o$.

\noindent\emph{Notation:} 
$\mathbf{e}$ denotes a unit direction, with $\mathbf{e}_x$, $\mathbf{e}_y$, $\mathbf{e}_z$ the
coordinate axes. $\mathrm{Rot}_x(\alpha)$ is a
rotation $\alpha$ about $\mathbf{e}_x$, likewise for $y$ and $z$. Images $I_i$, depth maps $D_i$ and masks $m$ are functions of the pixel, so
$D_i(\mathbf{x})$ is the depth at pixel $\mathbf{x} = [u,v]$. A point $\mathbf{p}^o$ in object's RestPose frame $o$ is transformed to camera frame $c$ using the pose as:
$\mathbf{p}^{c} = \mathbf{R}_{co}\mathbf{p}^{o} + \mathbf{t}_{c o}$.
Our RestPose canonicalization is the function
$F(\mathcal{M})$ of a mesh $\mathcal{M}$. 
The overview of RestPose's procedure is:
\begin{enumerate}[label=(\roman*),itemsep=1pt,topsep=2pt]
    \item {Generate} a mesh
                    $\mathcal{M} = G(I_{\text{ref}}, m_{\text{ref}}, \mathbf{K}\,[,D_{\text{ref}}])$
          from a reference view, using an image-to-3D generator~\cite{sam3d}, $G$; depth-conditioned only in the RGB-D setting.
    \item Compute the object's rest pose and {canonicalize} $\mathbf{R}_{on} = F(\mathcal{M})$ \cref{sec:restpose,sec:construct};
    \item {Render} $\omega$ templates from the canonical mesh $\mathbf{R}_{on}\mathcal{M}$;

    \item {Match} query to templates and {solve} a similarity transform
          (\cref{sec:reg});
    \item {Label} the \emph{flip} of every frame jointly and post hoc, from the per-frame
          outputs of (iv) alone (\cref{sec:gauge}). 
\end{enumerate}

Steps (i), (ii) are onboarding, which run once per object. Steps (iii), (iv) are inference, once per detection. Step (v) is post hoc and multi-view. 
The overview is illustrated in~\cref{fig:method}.

\subsection{RestPose Formulation}
\label{sec:restpose}

An object at rest on a plane lies on a facet of its convex hull. We assume a uniform
{surface} density, because without any prior on the generated mesh, this is a straightforward computable physical mass model
 - so the center of mass is the area-weighted facet centroid:
\begin{equation}
        \bar{\mathbf{g}} \;=\; \sum\nolimits_f a_f\, \mathbf{g}_f \;\Big/ \sum\nolimits_f a_f ,
    \label{eq:com}
\end{equation}
where $a_f$ and $\mathbf{g}_f$ are the area and the centroid of facet $f$. We first define a stable set of poses by selecting a resting hull facet $f$ with outward
normal $\mathbf{n}_f$ with $\bar{\mathbf{g}}$ projecting inside it. So the stable set and up-axes of each $f$ can be written as:
\begin{equation}
    \mathcal{S}(\mathcal{M}) = \big\{\, f \subset \partial\,\mathrm{conv}(\mathcal{M})\;:\; \Pi_f(\bar{\mathbf{g}}) \in f \,\big\},
    \quad \mathbf{u}_f = -\mathbf{n}_f ,
    \label{eq:stable}
\end{equation}
with $\Pi_f$ the orthogonal projection onto $f$. For each resting facet $f$ we compute the probability $P_f$, that the object settles on $f$ when dropped from a uniformly random orientation:
\begin{equation}
        P_f = \sum_{f' \,\rightsquigarrow\, f} \frac{\Omega_{f'}}{4\pi},
    \quad
    \Omega_{f'} = 4\arctan\sqrt{\tan\tfrac{\eta}{2} \prod_{i=1}^{3}\tan\tfrac{\eta-\beta_i}{2}} ,
    \label{eq:prob}
\end{equation}
where $f' \rightsquigarrow f$ means that an object first landing on facet $f'$
comes to rest on $f$. $\beta_i$ are the angles between the unit vectors from $\bar{\mathbf{g}}$ to the three vertices of $f'$,
and $\eta = \tfrac{1}{2}\sum_i \beta_i$. Curved sides discretise into many near-parallel facets, so we merge resting
facets into \emph{families} whenever $\langle \mathbf{u}_f, \mathbf{u}_{f'}\rangle > \cos\gamma$. A family has a
candidate up-axis $\mathbf{u}_{\mathrm{fam}}$ with probability
$P_{\mathrm{fam}} = \sum_{f \in \mathrm{fam}} P_f$, and $\mathcal{C}(\mathcal{M})$ is the set of candidate families.

\subsection{Defining a Unique RestPose }
\label{sec:construct}

\begin{figure}
    \centering
    \includegraphics[width=\linewidth]{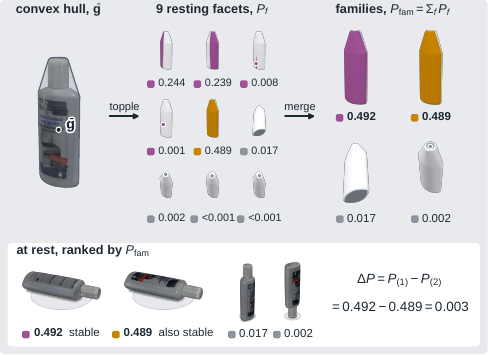}
    \caption{\textbf{Physics under-determines the canonical.} The rest-pose construction on an
        example bottle {mesh}. \emph{Top}: the convex hull and the centre of mass $\bar{\mathbf{g}}$, then the
    resting facets with their probabilities $P_f$, then the families they merge into.
    \emph{Bottom}: the bottle at rest in each family. Lying on its
    front and on its back differ by only $\Delta P = 0.003$, so
    stability alone cannot uniquely decide on a frame. 
     }
    \label{fig:underdet}
\end{figure}

The physics-based rest pose formulation of~\cref{sec:restpose} constrains the set of possible poses but does not determine it uniquely, i.e. an object can have multiple stable poses (\cref{fig:underdet}).
The figure also demonstrates that for some objects the difference between the highly probable rest poses is minuscule, making it hard to pick a single stable family.
Therefore, the RestPose formulation we have until now, narrows the frame to a small set of candidates and cannot choose among them, which is the problem we solve further in this subsection.

\paragraph{Breaking the tie with up-axis selection}
Let $\mathbf{e}_y$, the $+y$ axis of the native frame $n$, be the generated mesh's own upright axis. Of the candidate families $\mathcal{C}(\mathcal{M})$, we keep the one family whose up-axis is most nearly aligned with $\mathbf{e}_y$, and let $\mathbf{R}_{un}$ be the rotation that aligns that axis with $\mathbf{e}_z$, into an up-aligned frame $u$. We can calculate them by:
\begin{equation}
    \mathrm{fam}^{+} = \arg\max_{\mathrm{fam} \in \mathcal{C}(\mathcal{M})}\;
    \langle \mathbf{u}_{\mathrm{fam}},\, \mathbf{e}_y \rangle ,
    \quad \mathbf{R}_{un}\, \mathbf{u}_{\mathrm{fam}^{+}} = \mathbf{e}_z .
    \label{eq:prior}
\end{equation}

\begin{figure}
    \centering
    \includegraphics[width=\linewidth]{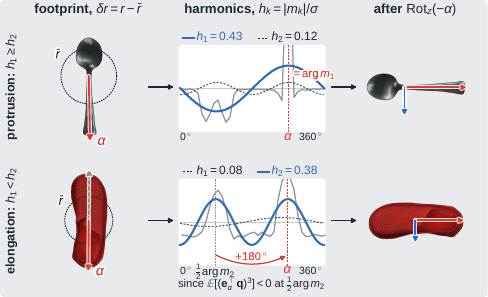}
    \caption{\textbf{The azimuth from the radial profile}, on two example meshes
    resting w.r.t their selected family, seen from above. \emph{Left}: $\delta r$ is each projected
    point's distance from the footprint's centre (the mean $\bar r$). \emph{Middle}:
    $\delta r$ summed in $10^\circ$ bins of $\theta$ (solid grey line) and its first two circular harmonics
    (\cref{eq:harm}), each drawn at amplitude $h_k/\pi$.
          \emph{Right}: the mesh after $\mathbf{R}_{hu} = \mathrm{Rot}_z(-\alpha)$, which puts the chosen direction on
    $+x$ (red). \emph{Top}: a spoon's handle is an off-centre protrusion: $h_1 > h_2$, so $\alpha = \arg m_1$. \emph{Bottom}: a shoe is elongated:
    $h_2 > h_1$, so $\tfrac{1}{2}\arg m_2$ fixes its long axis.
    }
    \label{fig:azimuth}
\end{figure}
\paragraph{Determining the Azimuth around the up-axis}
After computing $\mathrm{fam}^{+}$ and $\mathbf{R}_{un}$ (\cref{eq:prior}), only one degree of freedom remains undetermined for the RestPose definition, the azimuth $\alpha$ (\cref{fig:azimuth}).
To determine $\alpha$ we start with the ground-plane projection of the mesh under $\mathbf{R}_{un}$. Writing each projected point in polar form $(r_i,\theta_i)$, with $\delta r_i = r_i - \bar r$ its deviation from the mean radius and $\sigma^2$ the variance of $\delta r$, the first two harmonics~\cite{harmonic} of the radial profile are:
\begin{equation}
    m_k = \mathbb{E}\!\left[\delta r\, e^{\mathrm{i}k\theta}\right], \quad
    h_k = |m_k| / \sigma , \quad k \in \{1,2\} .
    \label{eq:harm}
\end{equation}
In Fourier shape description~\cite{kuhl1982}, $h_1$ responds to an off-centre protrusion, such as a handle, a spout
or a cap, and $h_2$ to an elongation, such as a box's long
side. In our case we let the stronger harmonic set the azimuth,
\begin{equation}
    \alpha =
    \begin{cases}
        \arg m_1, & h_1 \ge h_2 \\[2pt]
        \tfrac{1}{2}\arg m_2, & h_1 < h_2
    \end{cases}
        \qquad \mathbf{R}_{hu} = \mathrm{Rot}_z(-\alpha) .
    \label{eq:azimuth}
\end{equation}
Here $\arg m_k$ is the phase of the $k$-th harmonic, which is an angle in the ground plane measured
from $+x$, so $\alpha$ is one of the $\theta$ of \cref{eq:harm}, and $\mathbf{R}_{hu}$ aligns the harmonics-indicated direction with $+x$, into a heading-fixed frame $h$. A protrusion points somewhere, so the first case of \cref{eq:azimuth} is
unambiguous. An elongation does not, so the second fixes $\alpha$ only modulo $180^\circ$. We break
that tie with the third moment of the projected points along the chosen direction, taking
$\alpha + 180^\circ$ whenever $\mathbb{E}\big[(\mathbf{e}_\alpha^{\top}\mathbf{q})^3\big] < 0$, with
$\mathbf{e}_\alpha = (\cos\alpha, \sin\alpha)^{\top}$ and $\mathbf{q}$ the centered ground-plane coordinates.
Which means, an odd statistic changes sign under the half turn, so this resolves the azimuth as a function of
shape alone.
After composing the two previous steps, the final canonicalized rotation $\mathbf{R}_{on}$ of the mesh $\mathcal{M}$, then becomes:
\begin{equation}
     F(\mathcal{M}) \;=\; \mathbf{R}_{on} \;=\; \mathbf{R}_{oh}\; \mathbf{R}_{hu}\; \mathbf{R}_{un} .
    \label{eq:rc}
\end{equation}
with $\mathbf{R}_{oh} = \mathrm{Rot}_x(-90^\circ)$ to express the RestPose frame $o$ with $z$-up, 
only for registration \& matching convenience.

\subsection{Estimating Pose with the Canonical Mesh}
\label{sec:reg}
We now move directly to the inference module of the RestPose pipeline (see \cref{fig:method}). 
To achieve this, $\omega$ templates are rendered from the canonical mesh $\mathbf{R}_{on}\mathcal{M}$. Dense reciprocal matching~\cite{mast3r} between the query crop and each template returns correspondences
$(\mathbf{p}^{o}_i, \mathbf{x}_i)$, with $\mathbf{p}^{o}_i$ a mesh point in the canonical frame $o$ and $\mathbf{x}_i$ a query pixel. Lifting the pixel $\mathbf{x}_i$ to 3D using the query's depth $D(\mathbf{x}_i)$ and its homogeneous form $\tilde{\mathbf{x}}_i = (\mathbf{x}_i^{\top}, 1)^{\top}$, we can estimate the pose in the query's camera $c$ by:
\begin{equation}
\begin{aligned}
    \mathbf{p}^{c}_i &= D(\mathbf{x}_i)\, \mathbf{K}^{-1} \tilde{\mathbf{x}}_i , \\[2pt]
    (\hat{s}, \hat{\mathbf{R}}_{co}, \hat{\mathbf{t}}_{co}) &= \arg\min_{s,\mathbf{R},\mathbf{t}} \sum_i \big\| s \mathbf{R}\, \mathbf{p}^{o}_i + \mathbf{t} - \mathbf{p}^{c}_i \big\|^2 .
\end{aligned}
    \label{eq:umeyama}
\end{equation}
We can solve it using Umeyama~\cite{umeyama} within RANSAC~\cite{ransac} and select the template with the most inliers.
However, we employ two branches: When this matcher branch returns too few inliers, the detection
falls back to 
an {ICP}~\cite{icp} branch.

$(\hat{\mathbf{R}}_{co}, \hat{\mathbf{t}}_{co})$ in \cref{eq:umeyama} have no metric scale, 
and the generated mesh carries an arbitrary scaling factor $\mu$, and the depth predictor has no metric scale, so the depth may carry an arbitrary scaling factor $\lambda$.
Substituting $\mathbf{p}^{o}_i \mapsto \mu\, \mathbf{p}^{o}_i$ parametrizes the minimizer to $(\hat{s}/\mu,\, \hat{\mathbf{R}}_{co},\, \hat{\mathbf{t}}_{co})$, and substituting $\mathbf{p}^{c}_i \mapsto \lambda\, \mathbf{p}^{c}_i$ parametrizes to $(\lambda\hat{s},\, \hat{\mathbf{R}}_{co},\, \lambda\hat{\mathbf{t}}_{co})$.

{Translation inherits the depth's scale} since $\mu$ affects only $\hat{s}$. The output is therefore metric exactly when the depth is. In the \textbf{RGB-D setting}
$\lambda = 1$, $\hat{\mathbf{t}}_{co}$ is metric, and absolute thresholds $n^\circ m\,$cm apply. In the
\textbf{RGB-only setting} the depth is a predictor's with no metric anchor, so $\lambda$ is unknown, and we can only perform scale-agnostic pose estimation in this setting.
Rotation is invariant either way, which is why the same canonicalization $F$ serves both settings.

\subsection{Multi-View Processing for the Residual Flip}
\label{sec:gauge}

\begin{figure}[t]
    \centering
    \includegraphics[width=\linewidth]{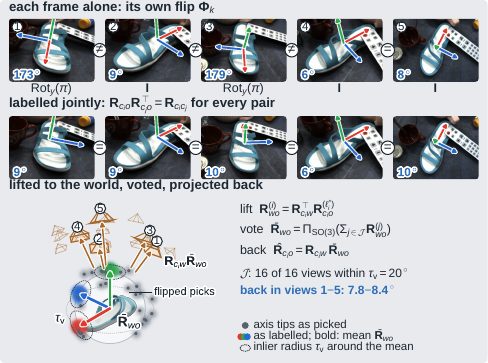}
    \caption{\textbf{Multi-view labelling and voting}
    with VGGT-predicted cameras. \emph{Top}: each frame's single-view pose carries its own flip
        $\boldsymbol{\Phi}_k$, and one view has no evidence to prefer one,
    so most pairs through them disagree with the cameras'
    relative rotation.
    \emph{Middle}: the labelling of \cref{eq:label} chooses the flips jointly.
    \emph{Bottom}: the $V{=}16$ object poses lifted to the world by the
    predicted cameras.
    Their chordal mean $\bar{\mathbf{R}}_{wo}$ is
    projected back into every view (\cref{eq:vote}).
    }
    \label{fig:gauge}
\end{figure}
In this section we try to address the \emph{near}-symmetry phenomenon that many objects show based on their appearance in the frame.
We first detect the geometric symmetry of the object and quotient it away:
\begin{equation}
        \angle_{\mathcal{A}}(\mathbf{R}) = \min_{\mathbf{a} \in \mathcal{A}(\mathcal{M}),\, \psi} \angle\big(\mathbf{R}\,\mathrm{Rot}(\mathbf{a},\psi)\big),
    \label{eq:contsym}
\end{equation}
where $\mathcal{A}(\mathcal{M})$ is the set of detected symmetry axes of the mesh $\mathcal{M}$, $\mathrm{Rot}(\mathbf{a},\psi)$ rotates by $\psi$ about the axis $\mathbf{a}$, $\angle_{\mathcal{A}}(\mathbf{R})$ the quotiented angle of a rotation $\mathbf{R}$, and $\angle(\mathbf{R}) = \arccos\big((\mathrm{tr}\,\mathbf{R} - 1)/2\big)$ the rotation angle.

What remains after the quotient is {near}-symmetry, where a rotation changes the shape too little for one view to see it, yet enough that the pose remains well defined. A near-symmetric object therefore still admits a discrete set of canonical \emph{flips}, defined by the Klein four-group $\boldsymbol{\Phi}_k \in \{\mathbf{I}, \mathrm{Rot}_x(\pi), \mathrm{Rot}_y(\pi), \mathrm{Rot}_z(\pi)\}$,
so a flipped pose $\mathbf{R}_{c_i o}\boldsymbol{\Phi}_k$ still maps $o$ into $c_i$. However, a single view cannot resolve them, so we make use of multiple views $V$ 
- for frames $i$ and $j$, the inter-frame rotation of a static object in the scene equals the camera's,
\begin{equation}
    \mathbf{R}_{c_i o}\,\mathbf{R}_{c_j o}^{\top} \;=\; \mathbf{R}_{c_i c_j} \;=\; \mathbf{R}_{c_i w}\,\mathbf{R}_{c_j w}^{\top},
    \label{eq:gauge}
\end{equation}
where $\mathbf{R}_{c_i o}$ is the object's rotation into the camera of frame $i$. The world-to-camera rotations
$\mathbf{R}_{c_i w}$ can be obtained from a multi-view geometry model~\cite{vggt}, and $\mathbf{R}_{c_i c_j}$ follows.

Each frame $i$ therefore carries a label $\ell_i = (b,k)$ selecting among its candidate poses
$\mathbf{R}^{(b,k)}_{c_i o} = \mathbf{R}^{(b)}_{c_i o} \boldsymbol{\Phi}_k$, with $\mathbf{R}^{(b)}_{c_i o}$ the rotation that branch $b$ estimates for frame
$i$. The candidates are the four flips from each of the two branches (Matcher or ICP in \cref{sec:reg}), predicted from that frame.
\cref{eq:gauge} scores a pair of labels by the quotiented angle of \cref{eq:contsym},
$c_{ij} = \angle_{\mathcal{A}}\big( \mathbf{R}^{(\ell_i)\top}_{c_i o}\, \mathbf{R}_{c_i c_j}\, \mathbf{R}^{(\ell_j)}_{c_j o} \big)$, which is zero
for correct poses. Therefore, the frame labelling can be obtained by:
\begin{equation}
    \boldsymbol{\ell}^{*} \;=\; \arg\min_{\boldsymbol{\ell}} \sum_{i<j}
    \min\!\big( c_{ij}(\ell_i,\ell_j),\, \tau \big) .
    \label{eq:label}
\end{equation}
The truncation angle $\tau$, set to $30^\circ$, removes the influence of outliers. We solve~\cref{eq:label} by iterated conditional modes, sweeping
$\ell_i \leftarrow \arg\min_{\ell} \sum_{j \neq i} \min(c_{ij}(\ell, \ell_j^{*}), \tau)$ over frames to convergence from several initialisations - to get the labelling of all frames, $\boldsymbol{\ell} = (\ell_1, \dots, \ell_V)$.

We can take further advantage of multiple views by voting over the labelled poses (see ~\cref{fig:gauge}). Each frame is
lifted to the shared world frame as $\mathbf{R}^{(i)}_{wo} = \mathbf{R}_{c_i w}^{\top} \mathbf{R}^{(\ell_i^{*})}_{c_i o}$, and the vote keeps the
largest inlier set $\mathcal{J}$ among
$\mathcal{J}_i = \{ j : \angle(\mathbf{R}^{(i)\top}_{wo} \mathbf{R}^{(j)}_{wo}) < \tau_{\mathrm{v}} \}$, with
$\tau_{\mathrm{v}} = 20^\circ$. Its consensus is:
\begin{equation}
    \bar{\mathbf{R}}_{wo} = \Pi_{\mathrm{SO}(3)}\Big(\textstyle\sum_{j \in \mathcal{J}} \mathbf{R}^{(j)}_{wo}\Big),
    \quad
    \bar{\mathbf{t}}_{wo} = \operatorname*{median}_{j \in \mathcal{J}} \mathbf{t}^{(j)}_{wo},
    \label{eq:vote}
\end{equation}
which is projected back to every query as $\hat{\mathbf{R}}_{c_i o} \leftarrow \mathbf{R}_{c_i w} \bar{\mathbf{R}}_{wo}$ (and $\bar{\mathbf{t}}_{wo}$ likewise). Where $\Pi_{\mathrm{SO}(3)}$ is the SVD projection onto the nearest rotation and $\mathbf{t}^{(j)}_{wo}$ is the frame's translation lifted by the same camera.
In the RGB-only setting, since we do not use metric-scale depth, only rotation is voted.

\section{Experiments}
\label{sec:exp}

\subsection{Implementation Details}
\label{sec:impl}

Onboarding generates one mesh per object from a single reference view using SAM-3D~\cite{sam3d}.
Canonicalization (\cref{sec:restpose,sec:construct}) runs on mesh geometry alone, with families merged at $\gamma=15^\circ$ and the azimuth taken from a uniform $8000$-point surface sample.
At inference we render the $\omega = 8$ templates on two rings, at elevation $0^\circ$ for azimuths $0/90/180/270$ and elevation $35^\circ$ for $45/135/225/315$, each at $384$ px. Dense reciprocal matching using MASt3R~\cite{mast3r} against each template feeds a $\mathrm{Sim}(3)$ Umeyama fit inside RANSAC, using $256$ iterations, and the template with most inliers is selected for pose estimation. Detections whose best template has $<30$ inliers fall back to a $24$-rotation initialized
{ICP}. 

In the RGB-D setting we also run {ICP} in parallel on every detection, and use it for all of an object's detections whenever the matcher's median inlier count over them is below $300$. Because on metric-scale depth {ICP} is more reliable, we can afford a higher inlier threshold than in the RGB-only setting.
There, the predicted depth of VGGT~\cite{vggt} is not metric-scale, so {ICP} serves only as the fallback when matching fails.
Multi-view stages use $V{=}16$ views per object per scene.
{No constant above is tuned per dataset}, and the same values run on all three benchmarks.

\subsection{Benchmarks and Protocols}
\label{sec:protocol}

Our estimates $\hat{\mathbf{R}}_{co}$ are expressed in the RestPose frame $o$, while a benchmark
annotates $\mathbf{R}^{*}_{ca}$ in the frame $a$ of its {CAD}s. We relate the two once per object by applying the same $F$ to the benchmark's {CAD} $\mathcal{M}_{\mathrm{CAD}}$, which gives the ground
truth in our frame,
\begin{equation}
    \mathbf{R}^{*}_{co} \;=\; \mathbf{R}^{*}_{ca}\,\mathbf{R}_{oa}^{\top},
    \qquad \mathbf{R}_{oa} = F(\mathcal{M}_{\mathrm{CAD}}) .
    \label{eq:align-main}
\end{equation}
The {CAD} is thus used only for scoring, after every prediction exists, and with the same function
and constants as onboarding. Any disagreement between $F$ on our generated mesh and on the {CAD} therefore caps the accuracy we can reach, which we measure in~\cref{sec:ablation}.
\begin{table}
  \centering\small
  \setlength{\tabcolsep}{4pt}
  \caption{Performance of RestPose (RGB-D variant) on \textbf{HouseCat6D} and comparison with other supervised baselines.}
  \begin{tabular}{@{}lcc@{}}
    \toprule
    \textbf{HouseCat6D} & $5^\circ2$cm & $10^\circ2$cm \\
    \midrule
    \multicolumn{3}{@{}l}{\emph{trained on the benchmark}}\\
    SecondPose~\cite{secondpose}         & 11.0 & 25.3 \\
    AG-Pose~\cite{agpose}                & 11.5 & 32.7 \\
    Sphere-Pose~\cite{spherepose}        & 19.3 & 40.9 \\
    Spot-Pose~\cite{spotpose}            & 23.8 & 52.3 \\
    GCE-Pose~\cite{gcepose}              & 24.8 & 55.4 \\
    OPT-Pose, single-view~\cite{optpose} & 28.0 & 53.0 \\
    OPT-Pose, multi-view~\cite{optpose}                 & 36.9 & 56.4 \\
    \midrule
    \multicolumn{3}{@{}l}{\emph{foundation model-based}}\\
    SAM-3D~\cite{sam3d}              & 20.5 & 43.0 \\
    \textbf{RestPose} (ours), single-view & 42.9 & 54.2 \\
    \textbf{RestPose} (ours), multi-view  & \textbf{55.2} & \textbf{65.9}\\
    \bottomrule
  \end{tabular}
  \label{tab:housecat6d}
\end{table}

\begin{table}
  \centering\small
  \setlength{\tabcolsep}{4pt}
  \caption{\textbf{REAL275}, scale-agnostic protocol from~\cite{optpose}. Here, depth at inference is predicted~\cite{vggt} and carries no
  metric scale (\cref{sec:reg}).
  }
  \begin{tabular}{@{}lcc@{}}
    \toprule
    \textbf{REAL275} & $5^\circ0.2d$ & $10^\circ0.2d$ \\
    \midrule
    \multicolumn{3}{@{}l}{\emph{trained on the benchmark}}\\
    DMSR~\cite{dmsr}                    & 15.1 & 25.6 \\
    LaPose~\cite{lapose}                & 15.7 & 37.4 \\
    GIVE-Pose~\cite{givepose}           & --   & 44.6 \\
    UniDet~\cite{unidet}                & 25.1 & 43.7 \\
    OPT-Pose~\cite{optpose}       & 24.3 & 44.1 \\
    \midrule
    \multicolumn{3}{@{}l}{\emph{foundation model-based}}\\
     SAM-3D~\cite{sam3d}               & 16.2 & 46.3 \\
    \textbf{RestPose} (ours), single-view & 26.4 & 47.3 \\
    \textbf{RestPose} (ours), multi-view  & \textbf{38.1} & \textbf{62.8} \\
    \bottomrule
  \end{tabular}
  \label{tab:real275}
\end{table}

\begin{table}
  \centering\small
  \setlength{\tabcolsep}{3.5pt}
   \caption{\textbf{Omni6DPose ROPE}.
  The benchmark ships
  no object meshes, so only relative pose protocol is possible for evaluation.
   Only the labelling stage is used here, since a consensus vote cannot be applied under a relative metric.
  }
  \begin{tabular}{@{}lcc@{}}
    \toprule
    \textbf{ROPE} (relative) & $<5^\circ$ & $<10^\circ$ \\
    \midrule
    OPT-Pose~\cite{optpose}                    & 34.2 & 55.3 \\
    \midrule
    \textbf{RestPose} (ours), single-view      & 34.6 & 51.8 \\
    \textbf{RestPose} (ours), multi-view       & \textbf{44.7} & \textbf{65.9} \\
    \bottomrule
  \end{tabular}
  \label{tab:rope}
\end{table}

We evaluate on the test sets of three benchmarks spanning $165$ categories. \textbf{HouseCat6D}~\cite{housecat6d} ($10$ categories) and \textbf{REAL275}~\cite{nocs} ($6$ categories) are scored against absolute annotations on their respective test scenes.
\textbf{Omni6DPose ROPE}~\cite{omni6dpose} ($149$ categories) has not publicly released its object meshes, so \cref{eq:align-main} cannot be formed. 
Therefore, on ROPE, we score relative poses between frame pairs - we infer on around $86$K frame pairs covering all categories and instances in the dataset.
\noindent\textbf{Metrics:}
HouseCat6D uses the standard combined thresholds $n^\circ m\,$cm. REAL275 uses the scale-agnostic protocol from~\cite{optpose}, which normalises translation error by the object's angular size. 
ROPE is scored by the relative residual
$\varepsilon_{ij} = \angle_{\text{sym}}(\mathbf{I},\, \mathbf{R}^{*\top}_{c_i a}\mathbf{R}_{c_i o}\mathbf{R}^{\top}_{c_j o}\mathbf{R}^{*}_{c_j a})$, quotiented by each instance's annotated symmetry order.

\subsection{Benchmark Evaluation}
\label{sec:sota}

\begin{figure*}[t]
    \centering
    \includegraphics[width=\linewidth]{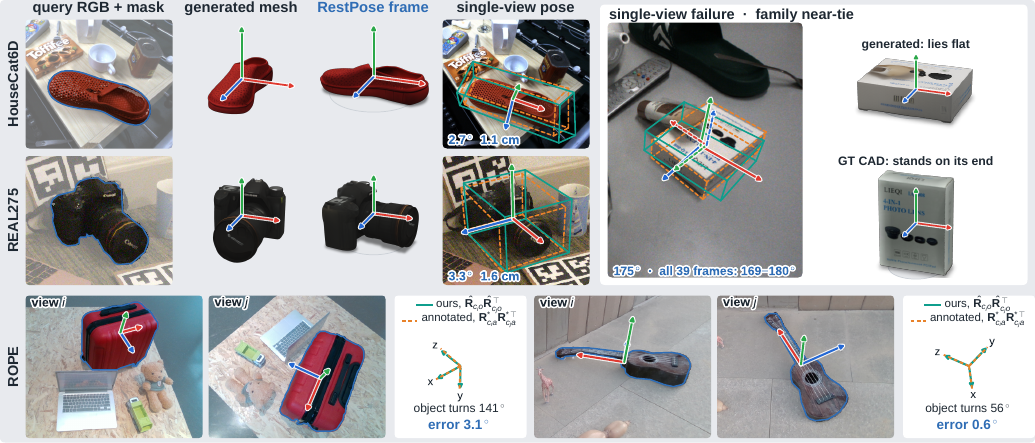}
    \caption{\textbf{Qualitative results.} HouseCat6D and REAL275 rows show the query object,
    the generated mesh in its native frame and in RestPose's frame, and the single-view pose,
    with the predicted 3D box in teal and the annotated box in dashed orange.
    \emph{Right}: the single-view
    failure that remains, the family near-tie of \cref{sec:mechanisms}. The generated box lies flat
    while the CAD stands on its end, each following its own source's upright, so their frames sit a
    face apart. The two boxes coincide, but the predicted axes differ from the annotation's
    by $169$ to $180^\circ$ in every one of the scene's frames. 
    \emph{Bottom}: Results on ROPE evaluated by relative pose. Each example shows two views,
    which stay on the same parts of the object as it
    turns, and compares our relative rotation with the annotated one, which
     coincide at these errors.
     }
    \label{fig:qualitative}
\end{figure*}

\Cref{tab:housecat6d,tab:real275,tab:rope} report the results on the three benchmarks and \cref{fig:qualitative}
shows representative qualitative results. 

\emph{HouseCat6D:}~\Cref{tab:housecat6d}, we lead at {every} combined threshold.
Against the baseline's~\cite{optpose} multi-view variant, we lead on both absolute pose metrics.
Even before any multi-view processing, our single-view result already shows improvements over prior methods.
\emph{REAL275.} \Cref{tab:real275} is produced without metric-scale, using our {RGB}-only reconstruction.
We lead decisively at both combined thresholds of the scale-agnostic metrics.
\emph{Omni6DPose ROPE.}~\Cref{tab:rope}, 
this is an important experiment to support our central idea - The benchmark contains $149$ categories compared to $10$ and $6$ categories in the other two benchmarks, which further challenges our canonicalization and pose-estimation.

Our single view variant is already on-par with the trained baseline method, and multi-view labelling further takes us ahead.
The consensus vote is not used here, since both frames of every pair come from one scene and would cancel the voting from their relative poses.

\subsection{Ablation Studies}
\label{sec:ablation}

\Cref{tab:ablation} ablates various stages: First, we look at how our multi-view labelling in~\cref{sec:gauge} brings
{coordination} among the frame predictions.
In random-labelling,
each frame draws its label uniformly from
the same eight candidates, the four flips in each of the two branches, over the same $V{=}16$
support.
Random-labelling scores well below the single view, because it changes
the flip over frames, more than three times as often as our multi-view labeller does, and leaves a median rotation error
above $120^\circ$.

Second, both multi-view stages add to the single view, and they repair different errors. The
labelling adds
performance by deciding the flip per frame, addressing near-symmetry.
The consensus vote then improves
further
by averaging the bimodal per-frame error that the labelling leaves.

The REAL275 block of \cref{tab:ablation} repeats the decomposition, where the ablations show similar performance gains. Here, we also add an articulation object prior for \emph{laptop} which improves the base score itself by a large margin. However, we reframe this ablation as a limitation in the further section.

\begin{table}
  \centering\small
  \setlength{\tabcolsep}{4pt}
    \caption{\textbf{Ablation} on HouseCat6D (top) and REAL275 (bottom), separating the stages of our proposed pipeline. 
  The lower REAL275 block adds the articulation prior described in~\cref{sec:mechanisms}.
  }
  \begin{tabular}{@{}lcccc@{}}
    \toprule
    & \multicolumn{2}{c}{no vote} & \multicolumn{2}{c}{$+$ vote} \\
    \cmidrule(lr){2-3}\cmidrule(l){4-5}
    \textbf{HouseCat6D} & $5^\circ2$cm & $10^\circ2$cm & $5^\circ2$cm & $10^\circ2$cm \\
    \midrule
    random labels (floor)                      & 21.2 & 28.1 & --    & --    \\
    single-view                                & 42.9 & 54.2 & 49.7 & 60.8 \\
    labelling for the flips                    & 47.9 & 58.3 & \textbf{55.2} & \textbf{65.9} \\
    \midrule[\heavyrulewidth]
    \textbf{REAL275} & $5^\circ0.2d$ & $10^\circ0.2d$ & $5^\circ0.2d$ & $10^\circ0.2d$ \\
    \midrule
    single-view                                & 26.4 & 47.3 & 35.0 & 58.1 \\
    labelling for the flips                    & 28.1 & 49.4 & \textbf{38.1} & \textbf{62.8} \\
    \midrule
    \multicolumn{5}{@{}l}{\emph{with the articulation prior}}\\
    single-view                                & 35.3 & 60.6 & 45.3 & 71.8 \\
    labelling for the flips                    & 36.7 & 62.8 & 46.4 & 74.9 \\
    \bottomrule
  \end{tabular}
  \label{tab:ablation}
\end{table}
\begin{table}
  \centering\small
  \setlength{\tabcolsep}{2pt}
    \caption{\textbf{Canonicalization agreement.}
  \emph{Top}: whether $F$ assigns the \emph{same}
  frame to different instances of one category, over a benchmark's released {CAD}s.
   \emph{Bottom}: whether $F$ returns the same frame on our
   generated mesh as on that object's reference {CAD}, giving us an empirical limit for our pose estimation. 
  }
  \begin{tabular}{@{}lrrrrrrr@{}}
    \toprule
    & & \multicolumn{2}{c}{full: $F$} & \multicolumn{2}{c}{up-only: gen.} & \multicolumn{2}{c}{up-only: phy.} \\
    \cmidrule(lr){3-4}\cmidrule(lr){5-6}\cmidrule(l){7-8}
     & \#inst. & $<15^\circ$ & $>90^\circ$ & $<15^\circ$ & $>90^\circ$ & $<15^\circ$ & $>90^\circ$ \\
    \midrule
    \multicolumn{8}{@{}l}{\emph{consistency across instances}}\\
    HouseCat6D    & 194  & \textbf{79.3} & 17.9 & \textbf{89.0} & 2.5 & 69.5 & 6.8 \\
    Omni6DPose   & 4216 & \textbf{73.8} & 17.0 & \textbf{93.5} & 1.0 & 79.2 & 8.1 \\
    \midrule
    \multicolumn{8}{@{}l}{\emph{agreement of generated vs.\ dataset's CAD}}\\
    HouseCat6D                     & 49   & \textbf{67.4} & 32.7 & \textbf{81.6} & 6.1 & 30.6 & 30.6 \\
    REAL275                         & 18   & \textbf{66.7} & 22.2 & \textbf{83.3} & 5.6 & 16.7 & 33.3 \\
    \bottomrule
  \end{tabular}
  \label{tab:ceiling}
\end{table}
\noindent\textbf{Effectiveness of canonicalization.}
The strongest generalization evidence of the canonicalization is provided here (\cref{tab:ceiling}). The test set canonicalization
residual is $67.4\%$ within $15^\circ$ on HouseCat6D and a similar $66.7\%$ on REAL275, on datasets that share no {CAD}s or annotation convention. The canonical also agrees with {itself} at closer rates on two different public {CAD} libraries ($79.3\%$ vs. $73.8\%$). 
While SAM-3D's own pose prediction is decent at 20.5\% $5^\circ2$cm (\cref{tab:housecat6d}) -
its generated meshes are reliably upright,  
against 
the physics-based up-axis.
So, we use generator's up-axis as a prior to break the tie between stable families.

\subsection{Limitations}
\label{sec:mechanisms}

We identify three main limitation areas for RestPose:

A \emph{family near-tie} arises where the generator's own upright is itself ambiguous
among near-equivalent faces, as on a box or a tube, so the generated mesh and the reference {CAD}
land a face apart (see \cref{fig:qualitative}). 
An \emph{azimuth near-tie}
arises for a body of revolution carrying one small feature, where the up-axis is recovered well and the
rotation about it (azimuth) is not: for example, on \emph{mug} the median up-axis error is $5.0^\circ$ while the full
$\mathrm{SO}(3)$ error is $17.0^\circ$.
\emph{Articulation}: Our rest pose formulation is defined for a rigid body. A hinged
object such as a laptop has inconsistent canonicalization due to its hinge angle determining the center of mass.

\section{Conclusion}

We introduced RestPose, a canonicalization approach that computes an object's canonical frame from
its own geometry differing from the current research where it comes from an annotation convention. 
We demonstrated that RestPose leads prior supervised methods on every
benchmark, despite using only frozen foundation models. Our ablations further show that the
computed canonical transfers across datasets without retuning.
while also exposing its limitation: near-symmetric and articulated objects.
Therefore, clear future works are formulations that resolve the near-symmetric
object problems (ideally without multi-view processing), and a derived canonical frame that also incorporates articulated objects. Moreover,
RestPose can also take advantage of future progress in RGB(-D)-based 3D generation and geometry foundation models.
We firmly believe that RestPose can pave the way for open object pose estimation, where objects of various categories can acquire consistent canonical frames without category-specific annotation or supervision.

\section*{Acknowledgements}
This research was funded by the European Union as part of the project: ShieldBOT (Grant Agreement 101235093).

{
    \small
    \bibliographystyle{ieeenat_fullname}
    \bibliography{main}
}

\clearpage

\setcounter{section}{0}\setcounter{equation}{0}\setcounter{table}{0}\setcounter{figure}{0}
\renewcommand{\thesection}{\Alph{section}}
\renewcommand{\theequation}{S\arabic{equation}}
\renewcommand{\thetable}{S\arabic{table}}
\renewcommand{\thefigure}{S\arabic{figure}}
\renewcommand{\theHsection}{supp.\Alph{section}}
\renewcommand{\theHequation}{supp.\arabic{equation}}
\renewcommand{\theHtable}{supp.\arabic{table}}
\renewcommand{\theHfigure}{supp.\arabic{figure}}

\section{Details of Evaluation Protocol}
\label{supp:eval}
\begin{figure*}[t]
  \centering
  \includegraphics[width=\linewidth]{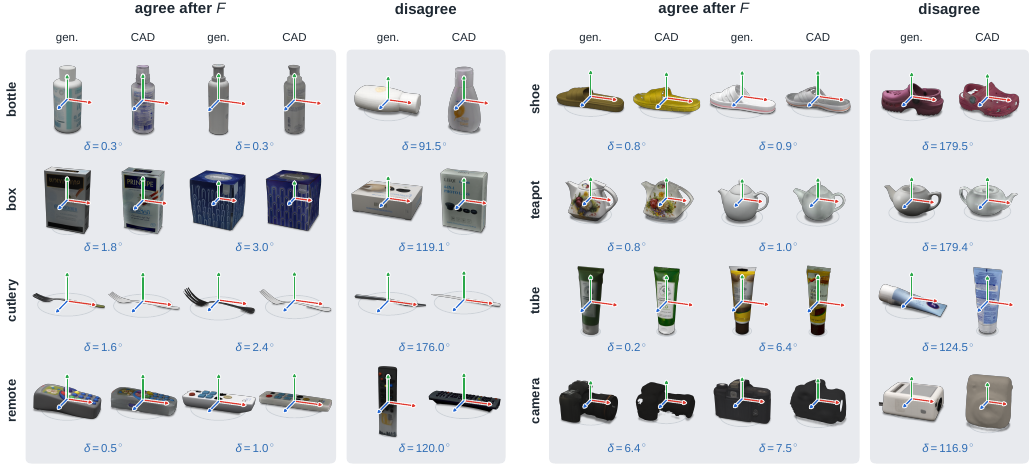}
  \caption{\textbf{$F$ on the generated mesh and on its GT CAD.} Each pair is one object, the mesh
  generated from a single image (gen.) and the object's GT {CAD}, both after $F$. Seven categories come from HouseCat6D, drawn from all its scenes since the
  method has no training, and the cameras from REAL275. Each row shows the two agreeing objects with the lowest $\delta$ and the disagreeing
  one with the highest. The failures are discrete, a different resting face for bottle, box, remote,
  tube and camera, and a $180^\circ$ turn for cutlery, shoe and teapot, the near-ties of
  \cref{sec:mechanisms}.}
  \label{supp:fig-gallery}
\end{figure*}

\paragraph{Angular error}
Rotation errors in the paper are a geodesic angle on $\mathrm{SO}(3)$,
\begin{equation}
    \angle(\mathbf{R}) = \arccos\!\Big(\tfrac{1}{2}\big(\mathrm{tr}\,\mathbf{R} - 1\big)\Big) \in [0,180^\circ],
    \label{supp:eq-angle}
\end{equation}
and for two rotations $\angle(\mathbf{R}, \mathbf{R}') = \angle(\mathbf{R}^{\top}\mathbf{R}')$, a metric on
$\mathrm{SO}(3)$ that is invariant under conjugation,
$\angle(\mathbf{Q}^{\top}\mathbf{R}\,\mathbf{Q}) = \angle(\mathbf{R})$. A subscript quotients it by a symmetry group
$\mathcal{G}$:
\begin{equation}
    \angle_{\mathcal{G}}(\mathbf{R}, \mathbf{R}') = \min_{\mathbf{S} \in \mathcal{G}}\, \angle\big(\mathbf{R}^{\top}\mathbf{R}'\mathbf{S}\big).
    \label{supp:eq-symangle}
\end{equation}
Each benchmark is scored under its own convention: HouseCat6D~\cite{housecat6d}
quotients the vertical axis of categories, bottle, can and glass, following VI-Net~\cite{vinet}. REAL275~\cite{nocs} does the same
for bottle, bowl, and can, and ROPE~\cite{omni6dpose} uses each instance's
annotated symmetry order.

\paragraph{Detected symmetry}
To keep the RestPose pipeline annotation-free, our canonicalization detects the symmetry of each generated mesh instance's geometry and quotients it instead of using the dataset's annotated symmetry. We measure self-similarity by the $95$th-percentile nearest-neighbour distance
$d(\cdot,\cdot)$ between a shape and a rotated copy of itself, and take
\begin{equation}
    \mathcal{G}(\mathcal{M}) = \big\{\, \mathbf{S} : d(\mathbf{S}\mathcal{M},\, \mathcal{M}) < \tau_{\mathrm{sym}} \,\big\},
    \label{supp:eq-symgroup}
\end{equation}
with $\tau_{\mathrm{sym}} = 0.02\,\mathrm{extent}(\mathcal{M})$, searching the $24$ cube rotations of the rest frame, the $n$-fold rotations about each of its axes,
and a $15^\circ$ sweep per axis that detects a continuous axis and yields the set
$\mathcal{A}(\mathcal{M})$ of \cref{eq:contsym}. 

\paragraph{The ROPE relative metric}
ROPE did not publicly release its object meshes, so there are no {CAD} absolute frames available. The benchmark is
scored on pairs of frames of one object. With $\mathbf{R}_{c_i o}$ our estimate in frame $i$ and
$\mathbf{R}^{*}_{c_i a}$ the annotation, the residual is
\begin{equation}
    \varepsilon_{ij} = \angle_{\mathrm{sym}}\big(\mathbf{I},\;
    \mathbf{R}^{*\top}_{c_i a} \mathbf{R}_{c_i o} \mathbf{R}_{c_j o}^{\top} \mathbf{R}^{*}_{c_j a}\big).
    \label{supp:eq-rope}
\end{equation}
Our estimates are measured in our canonical and the annotations in the benchmark's, which differ by a
per-instance offset $\mathbf{R}_{ao}$. If our estimate is exact up to that offset,
$\mathbf{R}_{c_i o} = \mathbf{R}^{*}_{c_i a}\mathbf{R}_{ao}$ in every frame, then
\begin{equation}
    \mathbf{R}^{*\top}_{c_i a} \big(\mathbf{R}^{*}_{c_i a} \mathbf{R}_{ao}\big)\big(\mathbf{R}^{*}_{c_j a} \mathbf{R}_{ao}\big)^{\top} \mathbf{R}^{*}_{c_j a}
    = \mathbf{R}_{ao} \mathbf{R}_{ao}^{\top} = \mathbf{I},
    \label{supp:eq-ropecancel}
\end{equation}
so a constant canonical offset cancels exactly and the metric measures pose alone. An offset that
varies between frames does not cancel, and that is precisely the discrete flip of \cref{sec:gauge}, so
the metric still sees the error canonicalization can cause. We report $\varepsilon_{ij}$ as class means
over the $149$ classes.

\paragraph{Why rotation thresholds rather than ADD(-S)}
Unseen-object methods often report ADD(-S) or a BOP average recall, but these metrics do not depend on the choice of canonical frame, by construction. If our frame differs from the benchmark's by a fixed $\mathbf{R}_{ao}$, so that our
model is $\mathbf{p}^{o} = \mathbf{R}_{ao}^{\top}\mathbf{p}^{a}$ and our estimate is
$\hat{\mathbf{R}}_{co} = \mathbf{R}^{*}_{ca}\mathbf{R}_{ao}$ with $\hat{\mathbf{t}}_{co} = \mathbf{t}^{*}_{ca}$. Therefore, each model point, transformed by our estimate, lands exactly where the annotated pose also transforms it in the following way:
\begin{equation}
    \hat{\mathbf{R}}_{co}\,\mathbf{p}^{o} + \hat{\mathbf{t}}_{co}
    = \mathbf{R}^{*}_{ca}\mathbf{R}_{ao}\mathbf{R}_{ao}^{\top}\mathbf{p}^{a} + \mathbf{t}^{*}_{ca}
    = \mathbf{R}^{*}_{ca}\mathbf{p}^{a} + \mathbf{t}^{*}_{ca},
    \label{supp:eq-addfree}
\end{equation}
so ADD and ADD-S are zero while the rotation error $\angle(\mathbf{R}_{ao})$ may be anything. This is what
allows reference-based methods  to report on shared splits without needing an annotated canonical frame
(\cref{tab:settings}). 

We therefore report each benchmark under the protocol of its category-level literature. Its rotation thresholds depend on the frame, so we first express our estimates in the benchmark's frame with \cref{eq:align-main}. Any disagreement between $F$ on our generated mesh and $F$ on the GT {CAD} then counts as rotation error even when the pose is right, which caps our scores (\cref{tab:ceiling}, bottom). We therefore count this error too rather than switch to a metric that ignores our computed canonical frame.

\paragraph{The SAM-3D baseline}
The SAM-3D~\cite{sam3d} rows of \cref{tab:housecat6d,tab:real275} read the generator's own pose output, a direct
regression of rotation, translation and scale with no correspondence step. SAM-3D runs on every
detection with the same inputs as our corresponding variant, and every run generates a new mesh. Each
generation is in its own canonical frame, so we register every generated mesh to the reference
{CAD} separately, with $24$ cube-rotation starts and {ICP}, which is more generous than our single alignment per object in \cref{eq:align-main}. 

Pose annotations for evaluation then follow each benchmark's convention.
With this expensive process of generating and registering on every frame, the results are decent but not close to the SOTA.
A regressed pose is
therefore useful across views only in a frame that stays fixed, which is the frame our approach computes. 

\section{On Canonicalization}
\label{supp:canon}

\paragraph{Why the flips are exactly four}
The candidate set $\boldsymbol{\Phi}_k$ of \cref{sec:gauge} is forced by the construction, and it is a
group. \Cref{eq:rc} fixes the frame in two stages, and each leaves a half turn ($180^\circ$) undetermined.
\Cref{eq:prior} selects one stable family, so the up-axis is inverted whenever the opposing family is
about as probable, the near-tie of \cref{sec:construct}. \Cref{eq:azimuth} then fixes the rotation about
that axis only modulo $180^\circ$, and the third moment that breaks this tie is an odd statistic, which
goes to zero when the shape is symmetric under the half turn, and is small when it is near-symmetric. The two
ambiguities are independent, one of the up-axis and one about it, so there are $2\times2=4$ candidates. With the frame $y$-up, as \cref{eq:rc} leaves it, the azimuth half turn is
$\mathrm{Rot}_y(\pi)=\mathrm{diag}(-1,1,-1)$, the inversion is $\mathrm{Rot}_x(\pi)=\mathrm{diag}(1,-1,-1)$,
and their composition is $\mathrm{Rot}_z(\pi)=\mathrm{diag}(-1,-1,1)$. These four rotations commute, are
each their own inverse and are closed under composition, so they form the Klein four-group
$\mathbb{Z}/2\times\mathbb{Z}/2$: $\boldsymbol{\Phi}_k \in \{\mathbf{I}, \mathrm{Rot}_x(\pi), \mathrm{Rot}_y(\pi), \mathrm{Rot}_z(\pi)\}$.

\paragraph{Measuring canonicalization agreement}
In \Cref{tab:ceiling}, its top block shows the extent of our canonicalization $F$  across the instances of the same class. Writing $\mathbf{R}_{o_k a} = F(\mathcal{M}_k)$ for the canonical of the $k$-th {CAD} of a
class and $\mathcal{G}_k$ for its detected symmetry group, we score every pair from both sides,
\begin{equation}
    \delta_{k\ell} = \min\big(\angle_{\mathcal{G}_k}(\mathbf{R}_{o_k a},\mathbf{R}_{o_\ell a}),\;
                              \angle_{\mathcal{G}_\ell}(\mathbf{R}_{o_\ell a},\mathbf{R}_{o_k a})\big),
    \label{supp:eq-ceilpair}
\end{equation}
take the best-connected instance as the consensus and report its deviations from all the others,
\begin{equation}
    k^{*} = \arg\max_{k} \big|\{\, \ell : 0 < \delta_{k\ell} < 15^\circ \,\}\big|,
    \qquad \{\, \delta_{k^{*}\ell} \,\}_{\ell \neq k^{*}},
    \label{supp:eq-ceilcons}
\end{equation}
averaged over classes. The bottom block of the table is the
same formulation, but once per onboarded object, between $F$ on our generated mesh and on that object's reference
{CAD},
\begin{equation}
    \delta = \angle_{\mathcal{G}}\big(F(\mathcal{M}_{\mathrm{gen}}),\, F(\mathcal{M}_{\mathrm{CAD}})\big).
    \label{supp:eq-achieved}
\end{equation}
The up-axis columns of the table score a single direction instead of a frame. Each generated mesh is registered to
its own reference {CAD}, and two up directions are mapped into the {CAD} frame, the generator's $+Y$ and
the up-axis of the most probable stable family alone. Their consistency across instances is measured as
above, and their agreement per object against the {CAD}'s own up-axis.

\paragraph{Tie-aware agreement}
\Cref{supp:eq-achieved} computes the agreement with the single best alignment, and reported in \cref{tab:ceiling}. However, a near-symmetric object looks almost the same after a rotation by $180^\circ$ about one of its axes, so the correct alignment of the generated mesh to the {CAD} and the alignment rotated by $180^\circ$ reach almost the same Chamfer distance, and the best alignment can land on the rotated one by a tiny margin. Counting every alignment within $10\%$ of the best Chamfer distance as equally good and keeping the smallest angle among them - the agreement rate on HouseCat6D rises within
$15^\circ$, with beyond $90^\circ$ decreasing, and REAL275 also shows the same behaviour in \cref{supp:tab-agree}. 
The agreement reported in \cref{tab:ceiling} is therefore conservative.
The native frame $n$ of the generated mesh is the generator's convention. Scored the same way against the dataset's frame, it agrees on far fewer objects than our canonical frame $o$.

\begin{table}[t]
  \centering\footnotesize
  \setlength{\tabcolsep}{2pt}
  \begin{tabular}{@{}lcccccccc@{}}
    \toprule
    & \multicolumn{4}{c}{$F$ frame ($o$)} & \multicolumn{4}{c}{SAM-3D native frame ($n$)} \\
    \cmidrule(lr){2-5}\cmidrule(l){6-9}
    & \multicolumn{2}{c}{strict} & \multicolumn{2}{c}{tie-aware} & \multicolumn{2}{c}{strict} & \multicolumn{2}{c}{tie-aware} \\
    \cmidrule(lr){2-3}\cmidrule(lr){4-5}\cmidrule(lr){6-7}\cmidrule(l){8-9}
    & $<\!15^\circ$ & $>\!90^\circ$ & $<\!15^\circ$ & $>\!90^\circ$ & $<\!15^\circ$ & $>\!90^\circ$ & $<\!15^\circ$ & $>\!90^\circ$ \\
    \midrule
    HouseCat6D & \textbf{67.4} & \textbf{32.7} & \textbf{79.6} & \textbf{18.4} & 26.5 & 49.0 & 32.7 & 34.7 \\
    REAL275    & \textbf{66.7} & \textbf{22.2} & \textbf{72.2} & \textbf{16.7} & 55.6 & 33.3 & 61.1 & 27.8 \\
    \bottomrule
  \end{tabular}
  \caption{\textbf{RestPose ($o$) vs SAM-3D ($n$) frames}. $F$ frame: $F$ on the
  generated mesh against $F$ on its {CAD}; its strict columns are from \cref{tab:ceiling}. Native frame: the generated mesh's own frame against the {CAD}'s. 
  Bold marks the better frame under the same scoring.}
  \label{supp:tab-agree}
\end{table}

\section{More Implementation Details}
\label{supp:impl}
\paragraph{Facet merging process}
\Cref{sec:construct} merges resting facets whose up-axes agree within $15^\circ$ and pools their
probabilities into $P_{\mathrm{fam}}$, but the normals of the facets are not averaged. Facets are visited in order of
decreasing $P_f$, the first unvisited one (the most probable of the merged facets) seeds a family, and every unvisited facet within the tolerance ($15^\circ$)
of that seed joins it. The family keeps the seed's up-axis and placement, the seed being by construction
its most probable member. We do this because: A mean direction over a $15^\circ$ cone is
in general not the normal of any hull facet, so the frame it induced would not itself be a resting
orientation, which is the property \cref{eq:stable} exists to guarantee.

\paragraph{Multi-view labelling}
Here, each object is processed post-hoc with its current frame and $15$ further frames of the same object, strided uniformly over
its track, so $V=16$. \Cref{eq:label} uses the truncation $\tau=30^\circ$ and is minimised by iterated
conditional modes from $128$ initialisations, each of which pins one frame's label and propagates from
it, keeping the labelling of lowest cost.

\paragraph{Support size}
\Cref{supp:tab-vsweep} varies $V$. Every support size improves on the single view on
all three benchmarks, and the full $V=16$ is best in every column. The reference method, OPT-Pose~\cite{optpose}
reports a four-view variant on HouseCat6D only, and our single-view row already exceeds it there at
$5^\circ2\,$cm, $42.9$ against $36.9$, so the comparison does not rest on multi-view processing.

\begin{table}[t]
  \centering\small
  \setlength{\tabcolsep}{3pt}
  \begin{tabular}{@{}lcccccc@{}}
    \toprule
    & \multicolumn{2}{c}{HouseCat6D} & \multicolumn{2}{c}{REAL275} & \multicolumn{2}{c}{ROPE} \\
    \cmidrule(lr){2-3}\cmidrule(lr){4-5}\cmidrule(l){6-7}
    $V$ & $5^\circ2$cm & $10^\circ2$cm & $5^\circ0.2d$ & $10^\circ0.2d$ & $<5^\circ$ & $<10^\circ$ \\
    \midrule
    $1$  & 42.9 & 54.2 & 26.4 & 47.3 & 34.6 & 51.8 \\
    $2$  & 45.9 & 57.9 & 29.6 & 53.8 & 35.5 & 52.7 \\
    $4$  & 47.6 & 58.8 & 32.1 & 61.0 & 37.7 & 55.7 \\
    $8$  & 50.2 & 60.4 & 32.7 & 61.3 & 39.9 & 58.8 \\
    $16$ & \textbf{55.2} & \textbf{65.9} & \textbf{38.1} & \textbf{62.8} & \textbf{44.7} & \textbf{65.9} \\
    \bottomrule
  \end{tabular}
  \caption{\textbf{Support size $V$} vs. performance of the multi-view stages on the three benchmarks (Omni6D ROPE only uses labelling stage without the consensus vote). $V=1$ is the single view, and $V=16$ is
  the multi-view row of \cref{tab:housecat6d,tab:real275,tab:rope}.}
  \label{supp:tab-vsweep}
\end{table}

\paragraph{Matcher choice}
We choose a pairwise matcher MASt3R~\cite{mast3r} to get the correspondences. Substituting a feed-forward multi-view model VGGT~\cite{vggt} for it, is worse on $9$ of $10$ HouseCat6D categories, with a median rotation error of
$39.5^\circ$ against $5.9^\circ$, and is $5\times$ slower. Therefore, we use VGGT only for its depth and the camera predictions in the multi-view processing case.

\end{document}